# Fusing Stylometric and Embedding Systems to Estimate Authorship Likelihood Ratios in Japanese

Praju Ghatpande, Satoru Tsuge, Shunichi Ishihara, Wataru Zaitsu, Mitsuyuki Inaba

**Abstract:** The likelihood ratio framework is widely recognized as the logically and legally sound basis for evidential analysis across forensic sciences, and its importance is increasingly acknowledged in analyses of authorship in textual evidence. To date, however, its application has been confined to English-language texts. Meanwhile, authorship attribution has traditionally relied on a diverse array of stylometric features, even as the rise of pre-trained large language models enables new contextual-embedding approaches. Combining these diverse approaches through fusion promises enhanced performance, yet it has not been applied to integrate stylometric-feature systems with embedding-based systems within the likelihood ratio paradigm. This study is the first to apply likelihood ratio-based forensic text comparison to Japanese digital texts—using ~1,000-character excerpts from blogs—to 1) evaluate system performance and likelihood ratio magnitudes and 2) assess the impact of fusing stylometric-feature systems with embedding-based systems. The results demonstrate that the fused system maintains excellent calibration while 1) increasing consistent-with-fact likelihood ratio magnitudes; 2) decreasing contrary-to-fact likelihood ratio magnitudes and 3) improving overall discriminability. The best-performing fusion achieved a log-likelihood-ratio cost of 0.32484, illustrating both the feasibility of likelihood ratio framework for Japanese and the benefits of fusion across heterogeneous systems.

## 1 INTRODUCTION

Authorship attribution—the task of inferring the author of texts with unknown provenance (Holmes 1994; Juola 2008; Koppel et al. 2012; Stamatatos 2009)—has a long history. Mendenhall's (Mendenhall 1887) landmark study is widely regarded as the first, demonstrating that each writer produces a distinctive distribution of word lengths. Subsequent seminal studies focused primarily on literary texts and articles (Mosteller and Wallace 1964; Yule 1939, 1944; Zipf 1932). With the rise of digital communication, research expanded to emails, blogs and social media (Stamatatos 2009), and more recently, forensic applications have come to the fore (Argamon 2018; Chaski 2005; Clark 2011; Grant 2007; Juola 2021; Mcmenamin 1993; Rocha et al. 2017).

Despite this expansion, the vast majority of authorship-attribution studies remain English-centric (Stamatatos 2009), with comparatively few investigations into non-European languages[1]—particularly those with fundamentally different writing systems, such as Chinese and Japanese. Japanese authorship attribution, however, is supported by a substantial body of fundamental research, including a comprehensive series of studies that evaluate a wide range of stylometric features and classification methods, thereby establishing robust approaches tailored to the specific characteristics of Japanese text (Jin 1994, 1996, 2002a, 2002b; Jin and Murakami 1993, 2007; Kanda and Jin 2024; Kanda et al. 2025; Zaitsu and Jin 2015, 2016, 2017, 2018a, 2018b).

Authorship attribution requires comparisons of multiple texts for which each text needs to be represented or modelled. Traditionally a common approach for text modelling is a bag-of-word (BoW) approach by which a text ($T_x$) is modelled via a numerical feature vector which consists of the counts of a given set of items (e.g., words): $T_x = \{x_1, x_2, x_3, \cdots, x_k\}$, where $x_k$ denotes the count of the $k^{th}$ word in the predefined vocabulary appearing in $T_x$. The items of a numerical vector, or features, does not have to be words, but can be anything which is considered to carry individuating information, and they can be a sequence of multiple items such as N-gram items where N > 1. There has been a variety of features having been proposed for the BoW approach, and it is reported that combining these features lead to an overall better performance (Zaitsu and Jin 2015, 2016, 2017).

More recently, the advent of pretrained large language models (LLMs) makes it possible to derive contextual embeddings that capture both local word usage and higher-order syntactic and semantic patterns, yielding rich representations of an author's stylistic characteristics. The representation based on embeddings have been a recent focus in the literature on deep learning-based authorship analysis, and have been found useful for a number of authorship analysis tasks (Boenninghoff et al. 2021; Boenninghoff et al. 2020; Hu et al. 2021; Ishihara et al. 2022; Shrestha et al. 2017; Zhu and Jurgens 2021). Even within the deep-learning-based systems, the performance of authorship attribution differs depending on the LLM used (Abbasi et al. 2022; Sun et al. 2023; Tyo et al. 2022), and the authorship information encoded in the resultant embedding vectors may be different.

Thus, authorship attribution can draw on diverse inputs—from traditional stylometric features to contextual embeddings produced by LLMs. Although a single representation may outperform others in isolation, combining the results of multiple representations often yields superior results. Combining the outputs of multiple systems is a well-established technique in natural language processing (NLP), commonly referred to as ensemble learning (Dietterich 2000; Zhou 2025). Ensemble learning reportedly boost performance in tasks such as text classification (Abburi et al.

[1] PAN <https://pan.webis.de/> is a leading initiative that organizes scientific events and shared tasks in digital text forensics and stylometry. It hosted authorship-attribution challenges in 2011, 2012, 2018 and 2019, targeting English, French, Italian and Spanish.

2023) and authorship analysis (Abbasi et al. 2022; Kanda and Jin 2024). In Japanese texts, for instance, document classification accuracy was improved by ensembling several LLM-based models trained on different data subsets (Chen et al. 2023). Similarly, the benefits of LLM-based ensembles for authorship attribution have been demonstrated (Kanda and Jin 2024). Furthermore, combining scores from LLM-based models with those from stylometric-feature systems yielded even higher attribution accuracy (Kanda et al. 2025).

With the rise of cybercrime—where malicious actors usually anonymize their identities online—the only link to perpetrators may be the incriminating texts they produce. This underscores the need for the authorship analysis in criminal investigations and court proceedings. In the legal cases in which the authorship of the incriminating texts is disputed, authorship analysis typically employ a verification paradigm, in which source-known and source-questioned texts are evaluated under two competing hypotheses—prosecution vs. defense (Ishihara 2023; Nini 2023; Nini et al. 2026). This differs subtly but importantly from non-legal applications of verification. Outside the courtroom, verification seeks to determine directly whether two texts share the same author. In legal proceedings, by contrast, the forensic expert's role is to assist the trier of fact by presenting objective evidential analyses; the ultimate determination of guilt remains solely with the judge or jury (Aitken and Taroni 2004; Champod and Meuwly 2000; Rose 2017). Any opinion by the expert that directly asserts authorship of the disputed text would exceed their remit.

Partly in response to the logical shortcomings of the traditional 'match' and 'identification' paradigms in forensic inference (Biedermann et al. 2008, 2016; Evett et al. 2017; Stoney 1991), there is growing consensus that the proper role of forensic experts is to evaluate the strength of the evidence—quantified as likelihood ratio (LR)—to assist the court's decision-making (Aitken et al. 2011; Aitken and Taroni 2004; Berger et al. 2011; Good 1991; Morrison 2012; Morrison et al. 2025; Robertson et al. 2016). The LR approach has been formally supported by several major professional and scientific bodies, including the Royal Statistical Society (Aitken et al. 2010), the European Network of Forensic Science Institutes (Willis et al. 2015), the American Statistical Association (Kafadar et al. 2019), and the Forensic Science Regulator for England and Wales (Forensic Science Regulator 2021). Its application has also been investigated across a wide range of forensic evidence types and inference problems, including voice comparison (Gonzalez-Rodriguez et al. 2007; Jessen 2025, 2026; Morrison et al. 2018), biological sex estimation from skeletal remains (Morrison et al. 2021b), hair comparison (Hoffmann 1991), toolmark analysis (Bunch and Wevers 2013), fingerprint comparison (Abraham et al. 2013; Alberink et al. 2014; Li et al. 2024; Morrison 2024b; Neumann et al. 2012), handwriting comparison (Bozza et al. 2008; Hepler et al. 2012; Johnson and Ommen 2022), evaporated gasoline residue analysis (Vergeer et al. 2014), firearm examination (Mattijssen et al. 2020), MDMA profiling (Bolck et al. 2015; Bolck et al. 2009), earmark comparison (Champod et al. 2001), footwear-mark comparison (Evett et al. 1998), and glass-fragment comparison (Curran 2003; Lambert et al. 2025; van Es et al. 2017). In the linguistic textual domain, and to emphasize this evidential focus rather than a definitive attribution, this study employs the term forensic text comparison (FTC). FTC captures the expert's task of comparing questioned and known texts and reporting the strength of evidence—rather than pronouncing on authorship—thereby aligning with best practices in LR-based forensic inference (Ishihara 2017, 2021; Ishihara et al. 2024).

Although the LR framework has been explored for FTC in English (Ishihara 2017, 2021, 2023; Nini 2023; Nini et al. 2026), it has, to the best of our knowledge, never been applied to non-English languages. Japanese—whose scripts, orthography and linguistic structure differ markedly from English—provides an ideal test case for assessing the framework's cross-language applicability.

Moreover, no prior work has examined ensemble learning or fusion (Brümmer and du Preez 2006; Morrison 2013) within the LR paradigm by combining traditional stylometric features with embeddings from LLMs.

Accordingly, this study focuses on Japanese digital texts to:

- Evaluate the performance of a LR-based FTC system and the magnitudes of its resulting LRs; and
- Assess the impact of fusing stylometric-feature systems with embedding-based systems.

More specifically, five stylometric-feature systems—based on Character bigrams, Function-word unigrams, Part-of-Speech (POS) bigrams, Comma bigrams and Character types (Section 4.1)—and two embedding-based systems—using word-level and character-level embeddings (Section 4.3)—are evaluated individually and in combination with fusion. To combine their outputs, logistic regression fusion is applied (Section 5.2). This study thus demonstrates the feasibility of the LR framework beyond English and the benefits of fusing heterogeneous systems, thereby addressing the research gap outlined above.[2]

### 1.1 Likelihood ratio framework

In forensic interpretation, the LR addresses a specific binary question framed by two competing propositions—commonly referred to as the prosecution and defense hypotheses (Morrison et al. 2021a; Morrison and Stoel 2014). Formally, the LR is defined as the ratio of the probability of the observed evidence under the prosecution hypothesis to its probability under the defense hypothesis (Aitken et al. 2010; Robertson et al. 2016). By Bayes' theorem, as shown in Equation (1), this LR converts prior odds into posterior odds, thereby updating the belief between the two hypotheses (Aitken and Taroni 2004):

$$\underbrace{\frac{p(H_p|E,I)}{p(H_d|E,I)}}_{Posterior\ odds} = \underbrace{\frac{p(H_p|I)}{p(H_d|I)}}_{Prior\ odds} \times \underbrace{\frac{p(E|H_p,I)}{p(E|H_d,I)}}_{Likelihood\ ratio} \quad (1)$$

where $E$ is the evidence, $H_p$ and $H_d$ are the prosecution and defense hypothesis, respectively, and $I$ is the relevant background information. Prior odds represent the trier of fact's beliefs about the prosecution and defense hypotheses before a new evidence (e.g., the textual evidence) is presented; posterior odds represent those beliefs after the evidence has been taken into account.

A LR is a quantitative statement of the strength of evidence, which is the ratio between the probability of the given evidence ($E$) assuming the prosecution hypothesis ($H_p$) being true and the probability of the same evidence assuming the defense hypothesis ($H_d$) being true. Conceptually, the numerator measures similarity—how similar or different the pieces of evidence under comparison—while the denominator measures typicality—how typical or atypical the same evidence is with respect to the relevant population. The LR, therefore, is the ratio of similarity to typicality.

Once a pair of texts under comparison are converted to vectors, the following LR calculation proceeds in two stages. First, a score is computed for each pair of texts under comparison. Second, this score is converted into a LR. This conversion is called calibration. Figure 1 illustrates this process.

[2] A link to the code used in the experiments is available from the corresponding author upon request.

Figure 1: Two stage process for calculating LRs. The section before the dashed line represents text pre-processing for vectorization. Blue boxes denote the score-calculation and calibration processes, while red boxes indicate the outputs of the preceding processes.

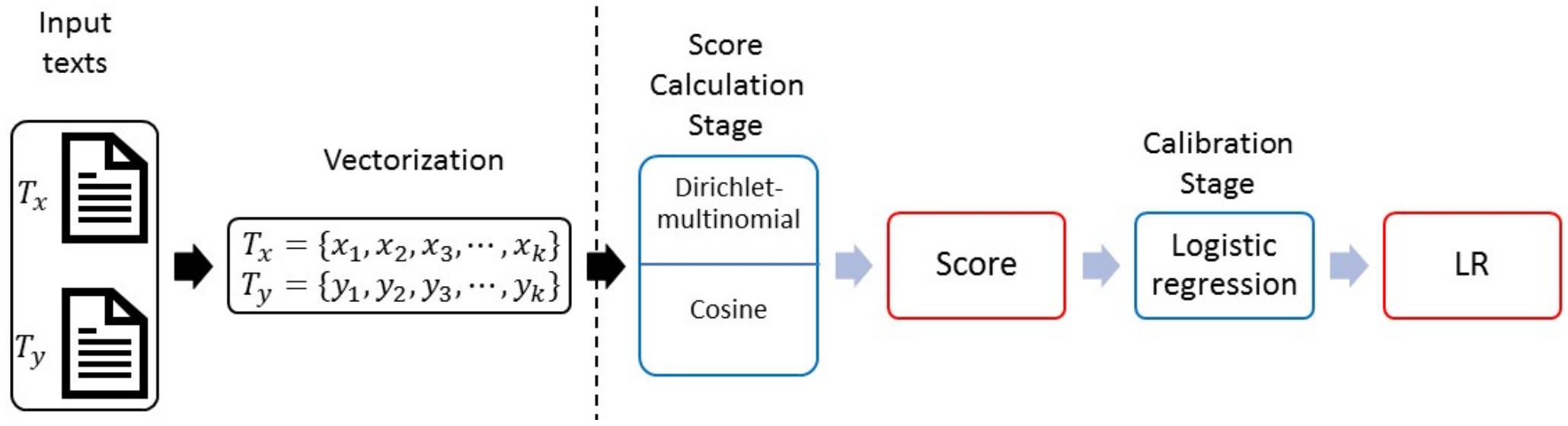

For the stylometric-feature systems—whose inputs are count-based—the Dirichlet-multinomial model was used to calculate these scores (refer to Section 5.1.1 for further details). For the embedding-based systems, scores were derived using cosine distance, the standard measure for comparing vector embeddings (Section 5.1.2). All scores were then converted to LRs through logistic regression calibration, which is the standard method for this purpose (Section 5.2). When multiple systems' scores are combined, logistic regression performs both fusion and calibration simultaneously, yielding a single set of LRs.

## 2 COMBINING MULTIPLE SYSTEMS

Combining the outputs of multiple systems is a well-established technique in NLP, commonly referred to as ensemble learning (Dietterich 2000; Zhou 2025). Ensemble learning reportedly boost performance in tasks such as text classification (Abburi et al. 2023) and authorship analysis (Fabien et al. 2020; Zamir et al. 2024). In Japanese NLP, for instance, document classification accuracy was improved by ensembling several BERT models trained on different data subsets (Tanaka et al. 2020). Similarly, the benefits of BERT-based ensembles for authorship attribution have been demonstrated (Kanda and Jin 2024). Furthermore, combining scores from BERT-based models with those from stylometric-feature systems yielded even higher attribution accuracy (Kanda et al. 2025).

One key advantage of the LR framework is its ability to combine evidential strengths from multiple sources—or from multiple features of the same source—into a single LR value (Aitken and Taroni 2004; Buckleton et al. 2018). This naturally aligns with the principles of ensemble learning. When individual LRs can be assumed independent, the overall LR is obtained by multiplying the component LRs—or, equivalently, by summing their log-LRs. If dependencies exist among the evidence sources, a fusion procedure that accounts for correlation is required (Kittler et al. 1998; Sergidou et al. 2024); logistic regression is a widely used method for learning an optimal weighted combination of LRs under such conditions (Brümmer and du Preez 2006; Morrison 2013). Logistic regression has been applied successfully to fuse LRs derived from multiple stylometric features in FTC studies (Ishihara 2014, 2023; Ishihara and Carne 2022). Accordingly, logistic regression is employed in this study to perform fusion and calibration simultaneously (see Section 5.2 for details).

## 3 DATABASE AND ITS PARTITIONING

A document database was compiled from Yahoo! Blog texts in the 'Balanced Corpus of Contemporary Written Japanese (BCCWJ)' (National Institute for Japanese Language and Linguistics 2021), covering posts dated between 26 April 2008 and 25 April 2009. Authors and their blog posts were selected as follows: [3]

1. Identify authors with at least two posts of 1,000 characters or more.
2. If an author contributed two or more eligible posts, randomly select two.
3. Manually review and exclude any posts unsuitable for FTC. [4]
4. Standardize each selected post to approximately 1,000 characters by truncating at the end of the last complete sentence before the limit is reached.

This procedure yielded 2,287 authors with two posts each, resulting in a database of 4,574 blog posts, each approximately 1,000 characters in length. An example blog post included in the database can be found in the appendix (A.1).

The entire 2,287 authors were divided exclusively into five batches as given below.

Table 1: The number of authors included in each batch.

| | Batch0 | Batch1 | Batch2 | Batch3 | Batch4 |
|---|---|---|---|---|---|
| Author numbers | 458 | 458 | 457 | 457 | 457 |

FTC experiments were carried out by swapping these batches as Training, Validation, Test and Calibration datasets in a cross-validation manner. As specified in able 2, there are five experiments.

able 2: Five-fold cross-validation experiments with 5 batches.

| | Training | Validation | Test | Calibration |
|---|---|---|---|---|
| Exp 1 | Batch0, Batch1 | Batch2 | Batch3 | Batch4 |
| Exp 2 | Batch1, Batch2 | Batch3 | Batch4 | Batch0 |
| Exp 3 | Batch2, Batch3 | Batch4 | Batch0 | Batch1 |
| Exp 4 | Batch3, Batch4 | Batch0 | Batch1 | Batch2 |
| Exp 5 | Batch4, Batch0 | Batch1 | Batch2 | Batch3 |

Note that Validation dataset is only applicable to the embedding-based systems.

## 4 SYSTEMS WITH STYLOMETRIC FEATURES AND EMBEDDINGS

This study employed two types of systems to calculate LRs: a discrete statistical model based on stylometric features and a model using pre-trained LLM embeddings. The stylometric-feature system incorporated five different feature sets, while the embedding-based system utilized two LLM variants. Detailed descriptions of both systems are provided in the following subsections.

### 4.1 System with stylometric features

The stylometric-feature system employs a Dirichlet-multinomial model to handle its count-based, discrete inputs. This approach has proven effective for estimating LRs from categorical data such as gunshot-residue evidence (Bolck and Stamouli 2017) and linguistic texts (Ishihara 2023; Ishihara et al. 2024). Five stylometric feature sets—each validated in prior Japanese authorship attribution research—were applied using this model (see Table 3).

---

[3] The pseudonyms were provided by the National Research Institute for Japanese Language and Linguistics under the research collaboration agreement.

[4] Some blogs consist solely of extensive lists of product names and prices.

Table 3: Stylometric features and their indices.

| Index | Stylometric features |
|---|---|
| 1 | Character bigram |
| 2 | Function-word unigram |
| 3 | POS bigram |
| 4 | Bigram of comma and the following character |
| 5 | Character types |

Each text was vectorized via a BoW representation using the stylometric features in Table 3. For the N-gram-based features (the first four entries in Table 3), the vector dimension was gradually incremented to see when the system starts converging in performance. In the following subsection, these stylometric features are described in detail after the morpho-syntactic characteristics of Japanese is briefly introduced.

## 4.2 Japanese language

Unlike English, Japanese does not use spaces at word boundaries. Thus, the first step of Japanese NLP is the application of a morphological analysis to a target sentence, splitting it into morphemes.

Table 4: Example of Japanese language processing—tokenizing/parsing a sentence into morphemes with their glosses. a) = An example sentence all written in Japanese; b) = English translation; c) = Tokenized sentence; d) = Romanized tokens; e) = Gloss; GEN = Genitive particle; LOC = Locative particle; NOM = Nominative particle; SYM. = Symbol; COP = Copula.

| | | | | | | | | | | | | | |
|---|---|---|---|---|---|---|---|---|---|---|---|---|---|
| a) | テレビの上にいる猫が、私の猫です。 | | | | | | | | | | | | |
| b) | The cat on the TV is my cat. | | | | | | | | | | | | |
| c) | テレビ | の | 上 | に | いる | 猫 | が | 、 | 私 | の | 猫 | です | 。 |
| d) | terebi | no | ue | ni | iru | neko | ga | 、 | watashi | no | neko | desu | 。 |
| e) | TV | GEN | top | LOC | exist | cat | NOM | SYM | I | GEN | cat | COP | SYM |

Japanese is a postpositional language in which particles mark the grammatical relationships between the nouns and the predicate of a sentence. In the case of the sentence given in Table 4, for example, the NOM particle (*ga*) marks that the preceding noun (*neko*) is the subject of the sentence. Sentence boundaries are marked by a small circle (。), and the comma (、) is placed whenever a brief pause aid readability and comprehensive; typically, between listed items, between clauses or phrases or after a particle with a noun phrase. However, there is no strict rule regarding the location of the comma; thus, the usage is somewhat stylistic, guided by clarity rather than rigid grammar.

### *4.2.1 Character bigrams*

Character-based N-grams are widely used in authorship attribution across languages (El and Kassou 2014; Rocha et al. 2017; Stamatatos 2009), and their effectiveness has been demonstrated for Japanese (Jin and Murakami 2007; Matsuura and Kanada 2000; Zaitsu and Jin 2018b). Texts were represented by the counts of frequent character bigrams extracted from the entire database, with vector dimensionality incrementally increased to 500. Table 5 lists the ten most frequent character bigrams, and the accompanying data file provides all 500.

Table 5: The ten most frequent character bigrams, with their frequencies and document frequencies in the entire database.

| Rank | Character Bigrams | Frequency | Doc-frequency |
|---|---|---|---|
| 1 | (‘っ’, ‘て’) | 36856 | 4517 |
| 2 | (‘し’, ‘た’) | 31629 | 4517 |
| 3 | (‘で’, ‘す’) | 28441 | 3908 |
| 4 | (‘っ’, ‘た’) | 27193 | 4437 |
| 5 | (‘し’, ‘て’) | 25200 | 4574 |
| 6 | (‘て’, ‘い’) | 23061 | 4250 |
| 7 | (‘す’, ‘。’) | 21838 | 3524 |
| 8 | (‘た’, ‘。’) | 19592 | 3781 |
| 9 | (‘な’, ‘い’) | 19435 | 4212 |
| 10 | (‘ま’, ‘す’) | 19323 | 3746 |

Note that the Character bigrams are represented in the form (‘Character1’, ‘Character2’).

### *4.2.2 Function word unigrams*

Mosteller and Wallace’s (Mosteller and Wallace 1964) seminal analysis of the Federalist Papers demonstrated that function-word frequencies can reliably attribute authorship. Consequently, function words have become nearly standard features in authorship attribution (Argamon 2018; García and Martín 2007; Grieve 2007; Stamatatos 2009), valued for their individual distinctiveness and resilience to topic variation, unlike content words. For Japanese, the frequencies of particles (e.g., GEN, LOC and NOM particles in Table 4), which are parts of function words, are known to perform well for authorship attribution (Jin 2002a, 2002b; Zaitsu and Jin 2018a). In the current study, other types of function words are also included in modelling texts. The functions words which are considered are: Pronouns, Conjunctions, Particles and Auxiliary verbs. As with character bigrams, texts were modelled using the counts of frequent function-word unigrams extracted from the entire database, with vector dimensionality incrementally increased to 500. These unigrams were grouped by their surface forms. Table 6 lists the ten most frequent function-word unigrams, including their frequencies and document frequencies, and the accompanying data file provides all 500.

Table 6: The ten most frequent function word (FW) unigrams, with their frequencies and document frequencies in the entire database.

| Rank | FW Unigrams | Frequency | Doc-frequency |
|---|---|---|---|
| 1 | ‘の’ | 132635 | 4574 |
| 2 | ‘に’ | 88117 | 4568 |
| 3 | ‘て’ | 85352 | 4558 |
| 4 | ‘た’ | 75674 | 4543 |
| 5 | ‘は’ | 74460 | 4564 |
| 6 | ‘が’ | 70846 | 4563 |
| 7 | ‘で’ | 65011 | 4563 |
| 8 | ‘と’ | 55247 | 4556 |
| 9 | ‘を’ | 54569 | 4529 |
| 10 | ‘も’ | 40394 | 4522 |

MeCab (mecab-python3 1.0.10),[5] a morphological analyser for Japanese text, was used in this study together with unidic[6] as a dictionary for tokenization and identifying the function words from the tokenized words.

*4.2.3 Part of speech bigrams*

MeCab was also used with unidic to obtain parts of speech (POS) information for each word. There are 16 top-level POS tags included in the POS system of unidic[7] (see Table 7), and some of them are further divided into sub-categories at the mid-level. All together there are 30 POS tags at the mid-level, and they are used as the basis of POS bigrams.

Table 7: 16 top-level POS tags in unidic.

| | | | |
|---|---|---|---|
| Noun | Pronoun | Adjectival | Pre-noun modifier |
| Adverb | Conjunction | Interjection | Verb |
| Adjective | Auxiliary verb | Particle | Prefix |
| Suffix | Symbol | Auxiliary symbol | Whitespace |

In this study, texts were modelled using the most frequent POS bigrams, with the number of features gradually increased to 500. The top ten POS bigrams are listed in Table 8 for reference.

Table 8: The ten most frequent POS bigrams, with their frequencies and document frequencies in the entire database.

| Rank | POS Bigrams | Frequency | Doc-frequency |
|---|---|---|---|
| 1 | (‘名詞-普通名詞’, ‘助詞-格助詞’) | 269544 | 4573 |
| 2 | (‘助詞-格助詞’, ‘名詞-普通名詞’) | 156246 | 4572 |
| 3 | (‘名詞-普通名詞’, ‘名詞-普通名詞’) | 138699 | 4574 |
| 4 | (‘助詞-格助詞’, ‘動詞-一般’) | 87831 | 4563 |
| 5 | (‘動詞-非自立可能’, ‘助動詞’) | 77265 | 4560 |
| 6 | (‘補助記号-読点’, ‘名詞-普通名詞’) | 59430 | 4387 |
| 7 | (‘動詞-一般’, ‘助動詞’) | 57184 | 4550 |
| 8 | (‘助動詞’, ‘補助記号-句点’) | 56849 | 4447 |
| 9 | (‘名詞-普通名詞’, ‘助詞-係助詞’) | 55125 | 4556 |
| 10 | (‘補助記号-句点’, ‘名詞-普通名詞’) | 53929 | 4486 |

Note that the POS bigrams are represented in the form (‘POS1’, ‘POS2’).

The full set of 500 is available in the accompanying data file.

*4.2.4 Bigram of comma and the following character*

As explained in Section 4.2, there is no strict rule regarding the location of the comma (、) in Japanese; thus, the usage is somewhat idiosyncratic, encoded with individuating information. The author-idiosyncratic nature of the comma usage has been exploited in Japanese authorship attribution, for example, focusing on the distributions of the words, characters and POS preceding the comma or their bigrams (Jin 1994, 1996; Jin and Murakami 1993; Zaitsu and Jin 2017). Texts were modelled using bigrams of the comma and its preceding character. Vector dimensionality was gradually increased to 200, with the counts of the most frequent comma-preceding character bigrams used as features (see the accompanying data file for the full set). The top ten bigrams are presented in Table 9.

[5] https://pypi.org/project/mecab-python3/
[6] https://github.com/polm/unidic-py
[7] https://hayashibe.jp/tr/mecab/dictionary/unidic/pos

Table 9: The ten most frequent bigrams of the comma and the preceding character, with their frequencies and document frequencies in the entire database.

| Rank | Comma Bigrams | Frequency | Doc-frequency |
|---|---|---|---|
| 1 | (‘は’, ‘、’) | 12285 | 3206 |
| 2 | (‘で’, ‘、’) | 9887 | 3347 |
| 3 | (‘が’, ‘、’) | 9726 | 3353 |
| 4 | (‘て’, ‘、’) | 9158 | 3231 |
| 5 | (‘ら’, ‘、’) | 4709 | 2391 |
| 6 | (‘も’, ‘、’) | 4484 | 2212 |
| 7 | (‘と’, ‘、’) | 4474 | 2359 |
| 8 | (‘に’, ‘、’) | 4298 | 2290 |
| 9 | (‘し’, ‘、’) | 3861 | 2236 |
| 10 | (‘り’, ‘、’) | 3689 | 2085 |

Note that the comma bigrams are represented in the form (‘character’, ‘、’).

*4.2.5 Character types*

As illustrated in Table 4, Japanese text employs multiple scripts (Martin 2004; Tsujimura 2007): hiragana and katakana, which are phonetic syllabaries, and kanji, which are logographic characters. Sentences often combine these scripts, though they may appear entirely in hiragana or katakana. Typically, grammatical elements are written in hiragana, lexical items in kanji, and katakana is reserved for foreign loanwords and sound-symbolic expressions. Authors may render the same word in hiragana, katakana or kanji to convey nuance or as a stylistic choice. Consequently, script selection can be idiosyncratic to the author and is leveraged in authorship attribution (Zaitsu and Jin 2015, 2016). In this study, beyond the basic hiragana, katakana and kanji categories, characters are further classified as detailed in Table 10.

Table 10: Character groups.

| Hiragana | Katakana | Kanji | Punctuation[a] | Digit | Others |
|---|---|---|---|---|---|

[a] ‘Punctuation’ is Japanese punctuation marks, and English punctuation marks are counted as ‘Others.’ English alphabets are also counted as ‘Others.’

### 4.3 System with embeddings

This study employed two embedding-based systems built on pre-trained RoBERTa (Robustly Optimized BERT Pretraining Approach) models: one using a word-level tokenizer and the other a character-level tokenizer—hereafter referred to as the word-embedding and character-embedding systems, respectively. RoBERTa (Liu et al. 2019) is an enhanced variant of BERT (Bidirectional Encoder Representations from Transformers) (Devlin et al. 2019) that offers superior performance as a pretrained model and has demonstrated efficacy in authorship attribution (Jones et al. 2022; Zhu and Jurgens 2021). The two embedding-based systems are summarized in Table 11.

Table 11: Two RoBERTa models and their indices.

| Index | Tokenization units |
|---|---|
| 6 | Word-embeddings |
| 7 | Character-embeddings |

Specifically, the word-embedding system utilized Megagon Labs' roberta-long-japanese model [8] —a longer-input variant of the Japanese RoBERTa model with its max_position_embeddings parameter increased to 1,282. The character-embedding system is based on the roberta-base-japanese-char-wwm model [9] developed by the Language Media Processing Lab of Kyoto University.

Following (Zhu and Jurgens 2021), texts were tokenized into word- or character-level units and then processed by RoBERTa models—configured as Siamese networks and fine-tuned on the training data (and validated on the validation data)—to extract their vector representations. In the experiments, the token limit for the word-level RoBERTa was set to 800 with a fine-tuning batch size of 8, and for the character-level RoBERTa the token limit was 500 with a batch size of 32. Both models employed the improved proxy-anchor loss function, with all other parameters adopted from (Zhu and Jurgens 2021).

## 5 SCORE CALCULATIONS, CALIBRATION AND FUSION

As summarized in Table 12, there are seven systems; the first five are stylometric-feature systems and the last two are embedding-based systems.

Table 12: Seven systems and their indices.

| Index | Stylometric features/Embedding types | System types |
|---|---|---|
| 1 | Character bigram | Stylometric |
| 2 | Function-word unigram | |
| 3 | POS bigram | |
| 4 | Comma bigram[a] | |
| 5 | Character types | |
| 6 | Word-embeddings | Embedding |
| 7 | Character-embeddings | |

[a] 'Comma bigram' = bigram of the comma and the preceding character.

For each system listed in Table 12, same-author (SA) and different-author (DA) scores were first calculated (see Section 5.1) and then converted to LRs, as illustrated in Figure 1, to assess individual performance. Note that the stylometric features 'Character bigram,' 'Function-word unigram,' and 'POS bigram' were evaluated with a maximum vector dimensionality of 500, while 'Comma bigram' was capped at 200. For these four systems (Systems 1–4), dimensionality was increased in the following sequence: 5, 10, 20, 30, 40, 50, 60; then by 20 up to 100; and finally by 25 increments up to each feature's maximum (500 for the first three, 200 for the last). This procedure reveals how performance varies as a function of feature-vector dimension, and when it begins to converge. The performance of each system is presented in Section 7.1.

The scores calculated for individual systems are fused to LRs as illustrated in Figure 2.

---

[8] https://huggingface.co/megagonlabs/roberta-long-japanese

[9] https://huggingface.co/ku-nlp/roberta-base-japanese-char-wwm

Figure 2: Fusion process.

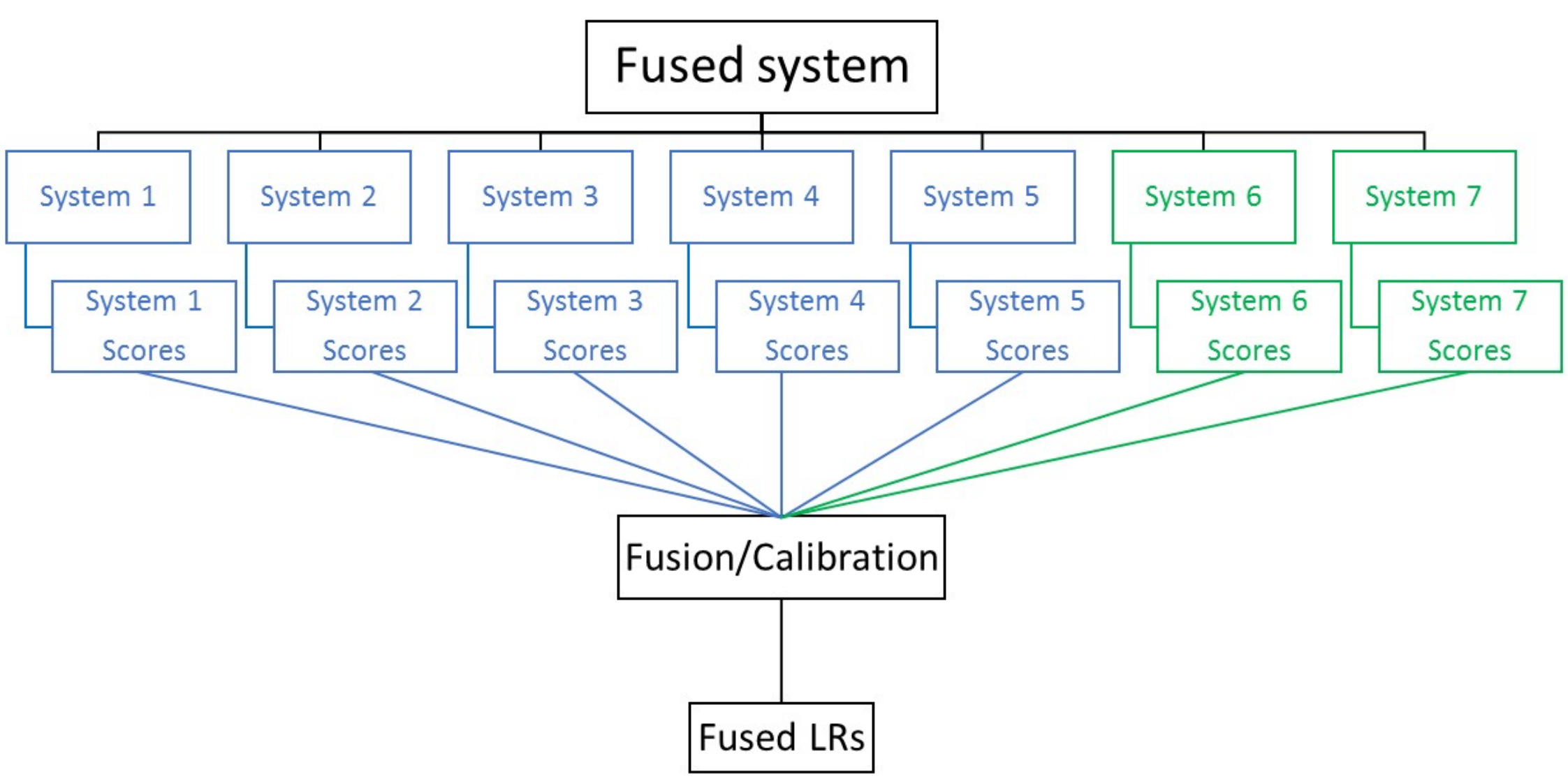

Blue = stylometric-feature systems and Green = embedding-based systems.

As can be seen from Figure 2, maximally seven sets of scores can be obtained from the seven systems; the first five are stylometric-feature systems and the last two are embedding-based systems. Were these scores independent, we could combine them by multiplying the linear scores or summing their $\log_{10}$ values. In practice, however, they exhibit correlations, so we employ a fusion technique to integrate them appropriately (see Section 5.2 for fusion).

All possible combinations of the seven systems from the two systems to seven systems are experienced for fusion. The numbers of possible combinations for fusion are summarized in Table 13.

Table 13: Numbers of combinations of fused systems.

| Number of fused systems | Numbers of combinations |
|---|---|
| 2 systems | 21 (= $\binom{7}{2}$) |
| 3 systems | 35 (= $\binom{7}{3}$) |
| 4 systems | 35 (= $\binom{7}{4}$) |
| 5 systems | 21 (= $\binom{7}{5}$) |
| 6 systems | 7 (= $\binom{7}{6}$) |
| 7 systems | 1 (=$\binom{7}{7}$) |

The fusion results for the stylometric-feature systems and the embedding-based systems are presented in Sections 7.2.1 and 7.2.2, respectively, while fusion across all seven systems appears in Section 7.2.3.

## 5.1 Score calculations

As noted in Section 1.1, the score-calculation methods differ between the stylometric-feature systems and the embedding-based systems. These are detailed in Sections 5.1.1 and 5.1.2, respectively.

*5.1.1 Score calculations: Stylometric-feature systems*

The stylometric features used in the first five systems (Table 12) are count-based and thus inherently discrete. Consequently, LRs were estimated with the Dirichlet-multinomial model—a method shown to be effective for categorical gunshot-residue evidence (Bolck and Stamouli 2017) and linguistic textual data (Ishihara 2023; Ishihara et al. 2024). Equation (2) gives the score for a pair of BoW-vectorized texts (e.g., $T_x = \{x_1, x_2, \cdots, x_k\}$, and $T_y = \{y_1, y_2, \cdots, y_k\}$), where $A = \{\alpha_1, \alpha_2, \cdots, \alpha_k\}$ is the Dirichlet parameter set and $B(\cdot)$ denotes the multinomial beta function.

$$score = \frac{B(A)B(A+T_x+T_y)}{B(A+T_x)B(A+T_y)} \qquad (2)$$

The Dirichlet parameters $A = \{\alpha_1, \alpha_2, \cdots, \alpha_k\}$ were estimated by maximum-likelihood using the reference dataset. A full derivation—from the LR formulation in Equation (1) to Equation (2)—is provided in (Ishihara 2023).

*5.1.2 Score calculations: Embedding-based systems*

In the embedding-based systems, once texts are vectorized by the fine-tuned RoBERTa models, cosine distance is computed between each pair of vectors (e.g., $T_x$ and $T_y$). See Equation (3) for the cosine distance, where $\langle \cdot, \cdot \rangle$ denotes the inner product.

$$score = cos(T_x, T_y) = \frac{\langle T_x, T_y \rangle}{\|T_x\| \, \|T_y\|} \qquad (3)$$

Two principal approaches exist for converting vectorized texts into scores. Equation (3) describes a similarity-only method (e.g., cosine distance), whereas Equation (2) describes a feature-based method (e.g. Dirichlet-multinomial scoring), considering both similarity and typicality. Each has its advantages and drawbacks, and some scholars have criticized similarity-only methods on theoretical grounds (Morrison and Enzinger 2018; Neumann and Ausdemore 2020), while others defend them for their practicality (Garton et al. 2020) or accept them when they demonstrably outperform relying on prior odds alone (Vergeer 2023). For in-depth critique of similarity-only scoring, see (Morrison and Enzinger 2018), and for a broader comparison of both approaches in FTC, see (Ishihara 2021; Ishihara and Carne 2022).

An alternative 'similarity-and-typicality score' approach—such as probabilistic linear discriminant analysis (PLDA) (Prince and Elder 2007)—requires multiple text samples per author (Ishihara et al. 2022), which this study does not provide (each comparison involves only two individual texts; see Section 3). Consequently, PLDA cannot be reliably trained here, and a similarity-only score (cosine distance) was adopted for practical reasons. Moreover, direct pairwise comparisons of single documents reflect a common and realistic FTC scenario.

## 5.2 Logistic regression calibration and fusion

A forensic evaluation system that produces LRs must be well calibrated—otherwise its outputs can mislead decision-makers (Gonzalez-Rodriguez et al. 2007; Meuwly et al. 2017; Morrison 2013, 2024a; Morrison et al. 2021a; Ramos and Gonzalez-Rodriguez 2013; Vergeer et al. 2020). In short, without good calibration, an LR loses its meaning as the weight of evidence, making calibration essential for any LR-based forensic inference system. Logistic regression is commonly used for this purpose (Brümmer and du Preez 2006; Morrison 2013), and also has been used in FTC (Ishihara 2023; Ishihara et al. 2024; Nini et al. 2026).

Calibration transforms raw SA and DA comparison scores into $\log_e$ LRs by applying a monotonic shift and scale that minimize classification error around a neutral decision threshold. As described by Morrison (2013: 181-182), in logistic regression calibration—performed in log-odds space—each score is first mapped to a posterior probability and then converted to posterior odds. Assuming equal prior odds (prior odds = 1), the logarithm of those posterior odds yields the natural-log LR ($\log_e$ LR). Finally, this $\log_e$ LR is converted to $\log_{10}$ LR for reporting. Logistic regression calibration is applied to each system's scores to assess its individual system performance.

Fusion extends calibration by combining multiple sets of scores—whether from different systems or different forensic sources—and mapping them to a single set of LRs. In logistic regression fusion, the $\log_e$ fused LR is computed as given in Equation (4).

$$log_e(\text{fused LR}) = \alpha + \beta_1 s_1 + \beta_2 s_2 + \beta_3 s_3 + \cdots + \beta_n s_n \quad (4)$$

where, $s_1, s_2, s_3, \cdots, s_n$ are the individual score sets (see Figures 1 and 2) and $\beta_1, \beta_2, \beta_3, \cdots, \beta_n$ are their corresponding scaling weights. The intercept $\alpha$ provides the necessary shift. All fusion weights ($\alpha$, $\beta_i$) are estimated from the SA and DA scores of the relevant calibration datasets.

## 6 LOG LIKELIHOOD RATIO COST AND TIPPET PLOT

The standard metric for LR-based inference systems—the log-LR cost ($C_{llr}$)—is used to assess the derived LRs (Brümmer and du Preez 2006; Morrison et al. 2021a; Morrison et al. 2020; Ramos et al. 2013). Detailed descriptions of $C_{llr}$ can be found in (Drygajlo et al. 2015; Gonzalez-Rodriguez et al. 2007; Meuwly et al. 2017; Morrison 2011; Morrison and Enzinger 2016; van Leeuwen and Brümmer 2007). See Equation (5) for its definition.

$$C_{llr} = \frac{1}{2}\left(\frac{1}{N_{SA}}\sum_{i}^{N_{SA}} log_2\left(1 + \frac{1}{LR_{SA_i}}\right) + \frac{1}{N_{DA}}\sum_{j}^{N_{DA}} log_2\left(1 + LR_{DA_j}\right)\right) \quad (5)$$

In Equation (5), $LR_{SA_i}$ and $LR_{SA_j}$ are linear LRs from the SA and DA comparisons, respectively. $N_{SA}$ and $N_{DA}$ are the number of SA and DA comparisons, respectively. $C_{llr}$ is a gradient metric, applying a penalty to each LR. The left and right terms of Equation (5) are for obtaining the average cost of all SA LRs and that of all DA LRs. At the end, the overall average cost is obtained between the average cost of all SA LRs and that of all DA LRs. The cost functions of $C_{llr}$ are illustrated in Figure 3.

Figure 3: The cost functions of $C_{llr}$.

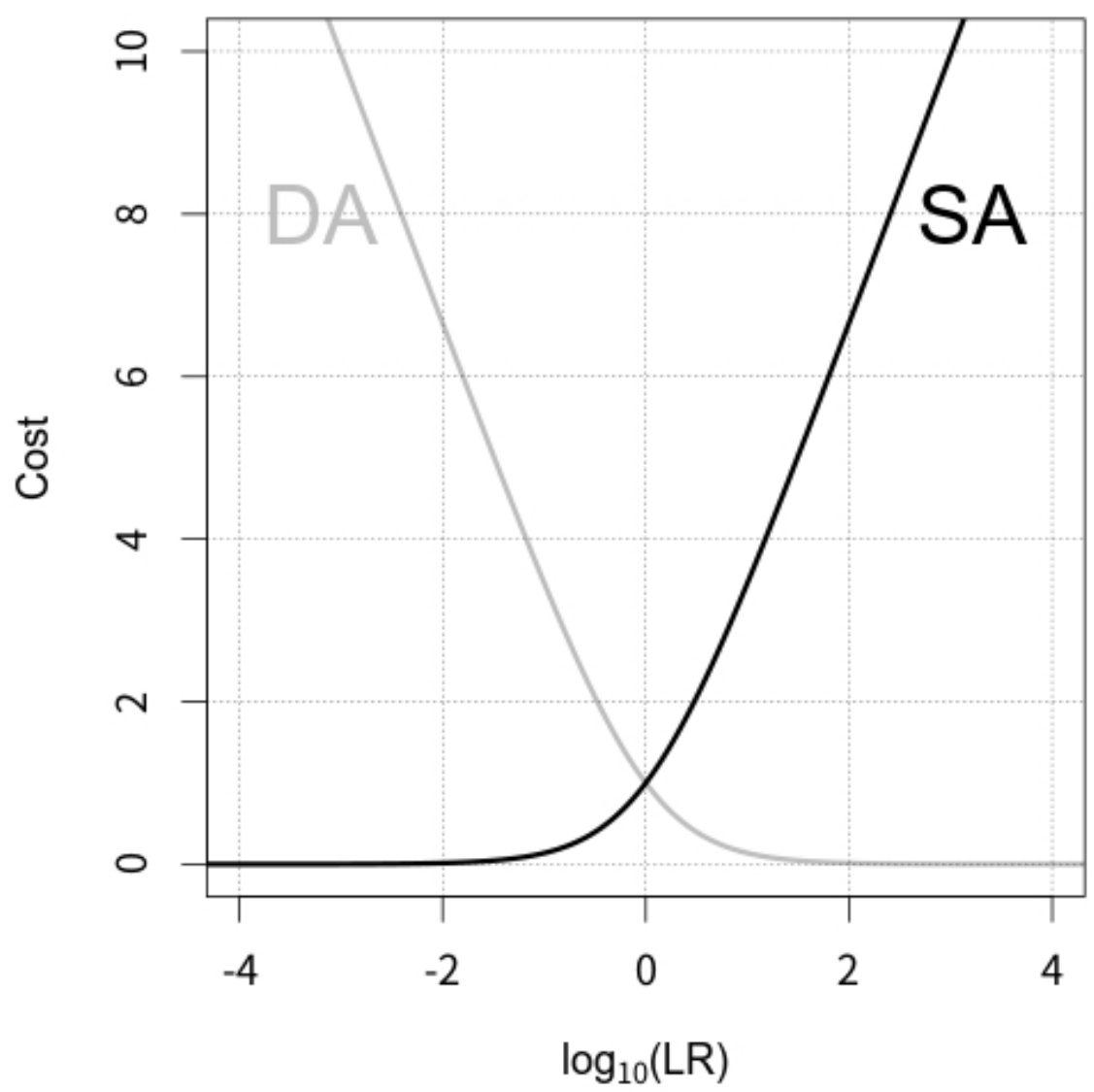

As shown in Figure 3, contrary-to-fact LRs (SS $\log_{10}$ LR < 0 or DS $\log_{10}$ LR > 0) incur substantial penalties in the $C_{llr}$ metric proportional to their magnitudes. The closer the $C_{llr}$ value is to zero, the better the system's overall performance. A $C_{llr}$ of 1 indicates no useful information—performance at chance level (Brümmer and du Preez 2006; van Lierop et al. 2024). $C_{llr}$ can be decomposed into two components—$C_{llr}^{min}$ (discrimination loss) and $C_{llr}^{cal}$ (calibration loss) (Ramos and Gonzalez-Rodriguez 2013)—but here only $C_{llr}$ and $C_{llr}^{cal}$ are reported, since $C_{llr} = C_{llr}^{min} + C_{llr}^{cal}$. Although the equal error rate (EER) is a familiar accuracy metric, it is categorical and therefore not strictly appropriate for continuous LR outputs. Nevertheless, EER values are provided for reference.

Tippett plots (Meuwly et al. 2017; Morrison et al. 2021a; Ramos et al. 2013) visualize both consistent-with-fact and contrary-to-fact LR magnitudes by displaying the empirical cumulative distributions for SA and DA comparisons as separate curves. Section 7.2.4 provides a detailed explanation of how to interpret these plots, where the derived LR magnitudes are presented.

## 7 EXPERIMENTAL RESULTS

This section presents the experimental results. Section 7.1 compares the performances of the seven individual systems (summarized in Table 12). Section 7.2 evaluates fusion in three stages: within the stylometric-feature systems (Section 7.2.1), within the embedding-based systems (Section 7.2.2), and across all seven systems—including every combination of two to six systems (Section 7.2.3; see Table 13 for counts of each fusion size). The effect of fusion on LR magnitudes is illustrated with Tippett plots in Section 7.2.4. Finally, Section 7.2.5 connects the results to the source texts by presenting the texts that yielded the highest SA LR and the lowest DA LR under the best-performing fusion.

### 7.1 Individual systems

The $C_{llr}$ values of the individual systems are plotted in Figure 3. The reader needs to be reminded that the feature dimensions of Systems 1, 2, 3 and 4 are gradually incremented to see when the system starts converging in performance; that is, the $C_{llr}$ values are plotted as the function of the feature dimension (or feature numbers) in Figure 4.

Figure 4: $C_{llr}$ values of the seven systems. Big circles indicate the best performance of Systems 1-5.

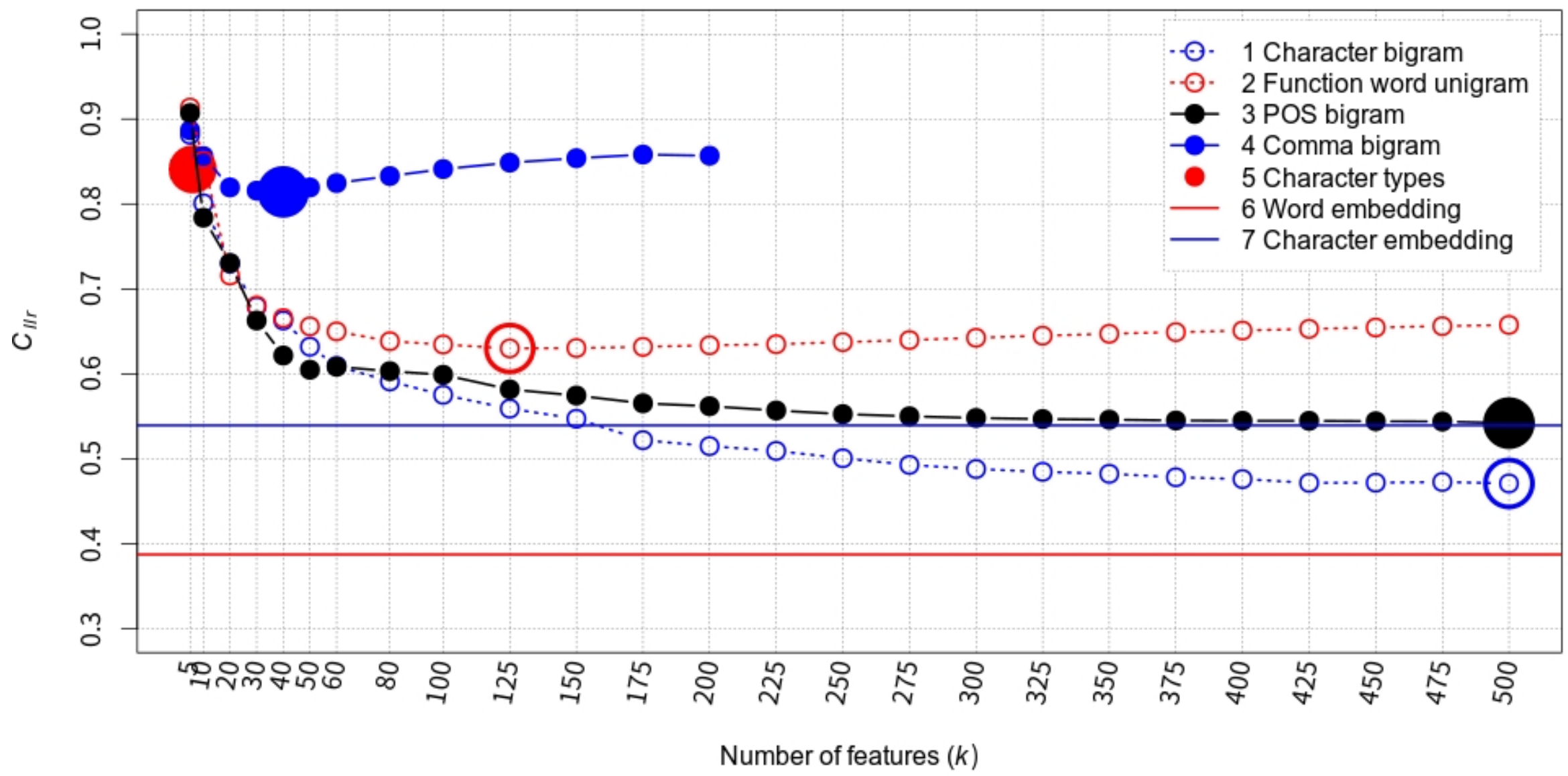

Since the embedding-based systems do not vary with vector dimensionality, their performance is represented by straight lines.

Systems 1, 2 and 3 employ feature vectors of up to 500 dimensions, whereas System 4 is limited to 200. In Figure 4, Systems 1 and 3 follow very similar $C_{llr}$ trajectories: a steep performance gain up to about 60 bigrams, then a slower but steady improvement to the 500-dimension limit, with marginal gains beyond 400. System 2 mirrors this initial trend—substantial improvement to 60 bigrams—but attains its optimum at 125 unigrams, after which performance gradually declines. System 4 shows a sharp rise in performance to 20 bigrams, peaks at 40, and then deteriorates as additional bigrams are added.

System 6, which uses word embeddings, achieved a $C_{llr}$ of 0.38757—outperforming both System 7 (character embeddings; $C_{llr}$ = 0.53951) and the best stylometric-feature system, System 1 (Character bigrams; $C_{llr}$ = 0.47115). System 7's $C_{llr}$ (0.53951) closely matches that of the POS-bigram system, System 3 ($C_{llr}$ = 0.54215). The performance gap between the two embedding-based systems may reflect the character-level RoBERTa's 500-token limit, which restricts vectorization to roughly half of each text sample.

Based on the optimal $C_{llr}$ values, the systems rank as follows: System 7 (0.38757) > System 1 (0.47115) > System 6 (0.53951) > System 3 (0.54215) > System 2 (0.63002) > System 4 (0.81432) > System 5 (0.84082).

### 7.2 Fused systems

This section first examines fusion at three levels—stylometric-feature systems (Section 7.2.1), embedding-based systems (Section 7.2.2), and all seven systems across two- to six-system combinations (Section 7.2.3)—and illustrates its impact on LR magnitudes via Tippett plots (Section 7.2.4). Section 7.2.5 then relates the findings to the data by showing the texts with the highest SA LR and the lowest DA LR under the best fusion.

*7.2.1 Fusion within stylometric-feature systems*

Five stylometric-feature systems—each built on one of the five stylometric features—were evaluated for fusion. All possible combinations of two to five systems were tested for fusion, and the best-performing fused system for each fusion size is listed in Table 14.

Table 14: Fused systems within stylometric-feature systems.

| Systems fused | Features | $C_{llr}$ | $C_{llr}^{cal}$ | $EER$ |
|---|---|---|---|---|
| 2 systems | 13 | 0.44280 | 0.04579 | 0.11978 |
| **3 systems** | **134** | **0.43964** | **0.04520** | **0.11841** |
| 4 systems | 1345 | 0.43972 | 0.04481 | 0.11903 |
| 5 systems | 12345 | 0.44065 | 0.04536 | 0.11921 |

Bold = best performing fused system.

Table 14 shows that the optimal $C_{llr}$ of 0.43964 is achieved by fusing the Character-bigram (System 1), Comma-bigram (System 3) and POS-bigram (System 4) systems. Improvements in $C_{llr}$ are marginal across different fusion sizes, and all $C_{llr}^{cal}$ values remain close to zero, confirming that every fused system is well calibrated. Notably, this best-performing fusion ($C_{llr}$ = 0.43964) outperforms the leading single-stylometic-feature system (Character-bigram; $C_{llr}$ = 0.47115), demonstrating the efficacy of fusion.

*7.2.2 Fusion within embedding-based systems*

Separately from the stylometric-based systems, the two embedding-based systems (Systems 6 and 7) were fused, and the metric values of the fused system are given in Table 15.

Table 15: Fused systems within embedding-based systems.

| Systems fused | Features | $C_{llr}$ | $C_{llr}^{cal}$ | $EER$ |
|---|---|---|---|---|
| 2 systems | 67 | 0.36186 | 0.03002 | 0.10446 |

Table 15 demonstrates that fusing the two embedding systems ($C_{llr}$ = 0.36186) yields greater improvement than the independent embedding-based systems, further confirming fusion's efficacy. Moreover, this embedding fusion outperforms the optimal fusion of the stylometric-feature systems ($C_{llr}$ = 0.43964).

*7.2.3 Fusion across stylometric-feature and embedding-based systems*

Sections 7.2.1 and 7.2.2 demonstrated that fusion was effective within both the stylometric-feature systems and the embedding-based systems. In this section, all possible combinations across the seven systems are evaluated for fusion. Table 16 lists the best-performing fused system for each fusion size.

Fusing System 1 (Character bigrams) with System 6 (Word embeddings)—thereby combining a stylometric-feature system and an embedding-based system—yields a $C_{llr}$ of 0.34603, outperforming the fusion of the two embedding-based systems alone ($C_{llr}$ = 0.36186). Incorporating System 7 (Character embeddings) further reduces $C_{llr}$ to 0.32907. Although adding more systems produces diminishing improvements, the fusion of Systems 1, 4, 5, 6 and 7 achieves the best performance overall ($C_{llr}$ = 0.32484; see Table 16). The $C_{llr}^{cal}$ value (0.03283) is very close to zero, showing the best-performing fused system is very well calibrated.

Table 16: Fused systems across machine-learning and deep-learning systems.

| Systems fused | Features | $C_{llr}$ | $C_{llr}^{cal}$ | $EER$ |
|---|---|---|---|---|
| 2 systems | 16 | 0.34603 | 0.03706 | 0.09716 |
| 3 systems | 167 | 0.32907 | 0.03406 | 0.09344 |
| 4 systems | 1467 | 0.32545 | 0.03463 | 0.09273 |
| **5 systems** | **14567** | **0.32484** | **0.03283** | **0.09288** |
| 6 systems | 124567 | 0.32579 | 0.03279 | 0.09341 |
| 7 systems | 1234567 | 0.32855 | 0.03555 | 0.09289 |

Bold = best performing fused system.

The $C_{llr}$ of the best fusion system has been added to Figure 4, and the corresponding plot is presented separately as Figure 5, enabling a clear visual comparison with the seven individual systems.

Figure 5: $C_{llr}$ values of the seven systems (Figure 4) with the $C_{llr}$ of the best fused system.

### *7.2.4 Magnitude of derived likelihood ratios*

To illustrate how fusing multiple systems affects the magnitude of the derived LRs, this section presents three Tippett plots in Figure 6 for:

- The best fusion among the stylometric-feature systems (Systems 1, 3 and 4): $C_{llr}$ = 0.43964
- The fusion of the two embedding-based systems (Systems 6 and 7): $C_{llr}$ = 0.36186
- The overall best fusion (Systems 1, 4, 5, 6 and 7): $C_{llr}$ = 0.32484

Figure 6: Tippett plots of the derived LRs plotted separately for the SA and DA $log_{10}$ LRs. Panel a) = the derived LRs fusing Systems 1, 3 and 4; Panel b) = the derived LRs fusing Systems 6 and 7; Panel c) = the derived LRs fusing Systems 1, 4, 5, 6 and 7.

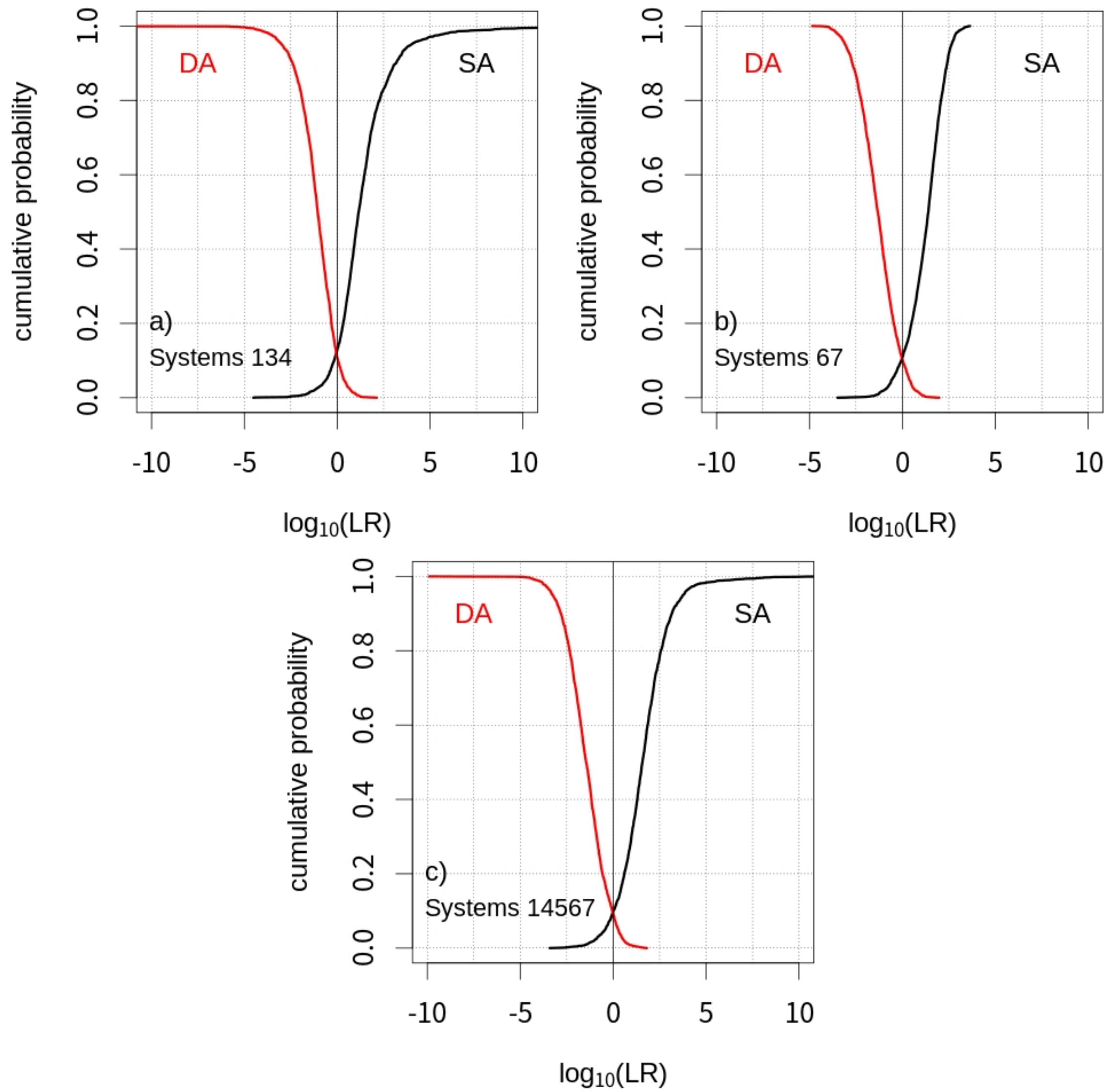

Note that, owing to the x-axis limits, both the SA and DA curves are truncated in Panel a), and the SA curve is likewise truncated in Panel c).

A Tippett plot depicts the magnitudes of LRs by showing the empirical cumulative distributions of $log_{10}$ LRs for SA and DA comparisons (Meuwly et al. 2017; Morrison and Enzinger 2016; Ramos et al. 2013). On the SA curve (which rises to the right), the y-axis value corresponding to a given value on the x-axis indicates the proportion of SA comparisons whose $log_{10}$ LR is ≤ that value. Conversely, on the DA curve (which rises to the left), the y-axis value corresponding to a given value on the x-axis shows the proportion of DA comparisons whose $log_{10}$ LR is ≥ that value.

Ideal performance appears as two broadly separated, relatively flat curves: the SA curve shifted towards large positive values and the DA curve towards large negative values. A perfectly calibrated system will have the curves intersect exactly at $log_{10}$ LR = 0. The Tippett plot simultaneously displays:

- Consistent-with-fact LRs—SA $\log_{10}$ LR > 0 or DA $\log_{10}$ LR < 0—where larger magnitudes indicate better performance; and
- Contrary-to-fact LRs—SA $\log_{10}$ LR < 0 or DA $\log_{10}$ LR > 0—where smaller magnitudes (i.e. closer to zero) denote better performance.

Thus, a flatter curve with wide separation and minimal tail-mass on the wrong side of zero signifies both good performances.

Figure 6's Tippett plots visually demonstrate that all three fused systems are well calibrated, as confirmed by their $C_{llr}^{cal}$ values being close to zero: 0.04520 (Panel a), 0.03002 (Panel b) and 0.03283 (Panel c). In terms of $C_{llr}$, the fused system of which the LRs are displayed in Panel c) ($C_{llr}$ = 0.32484) is the best system than those systems displayed in Panels a) ($C_{llr}$ = 0.43964) and b) ($C_{llr}$ = 0.36186). The LRs displayed in Panel c) well account for the superiority of the system in performance (confirmed by $C_{llr}$) in comparison to the other systems: Panels a) and b). For example, when examining consistent-with-fact LRs, the SA and DA curves in Panel c) lie further from the neutral point ($\log_{10}$ LR = 0) than those in Panel b). Although Panel a) attains the greatest consistent-with-fact SA and DA LR values (SA = 62.20686; DA = −18.74429 vs. 28.03885 and –9.90240 in Panel c), Panel c) displays a greater overall magnitude in the consistent-with-fact LRs—mean SA = 1.88653 and mean DA = –1.65704 compared with 1.73760 and –1.33218 for Panel a).

For the contrary-to-fact LRs, the fusion of Systems 1, 4, 5, 6 and 7 (Panel c) reduces the extremes: its maximum contrary-to-fact SA and DA LRs (–2.3962 and 1.8112) are smaller than those in Panel a) (–4.5110 and 2.1291) and Panel b) (–3.4896 and 1.9800). Although these improvements are not immediately obvious from Figure 2's Tippett plots, this same fusion also achieves the lowest EER (0.09288) among the three systems.

*7.2.5 Example texts*

To link the experimental results with the source texts, Table 17 presents an excerpt from one of the two texts in the SA comparison that yielded the highest consistent-with-fact SA LR and excerpts from the two texts in the DA comparison that yielded the lowest consistent-with-fact DA LR, under the best-performing fused system.

Table 17: Experts from the texts of sample IDs OY14_16071 (Text a) and OY07_01437 (Text b).

| Text a) |
| --- |
| ５０００万位でも、１億でも、くれるなら、、良い、、、、、何かを、、、企んでいる、、、、顔だな、、、、これに、比べると、豆タヌキは、、、安心して、見られるよ、、、、私の同級生で、ますぞえ大臣から、、厚生年金の手紙が来てるのは、、、、私だけだな、、、私は、ますぞえ君から、、選ばれた、特別な人民、、、、なのだよ、、、、ははははは、、、ますます、、ますぞえが、、気に入った、、、、、、豆タヌキは、変化なしだ、、、、生きている言葉は無しだ、、、、、、へへへ、、、、、と、同じ事を、四国でやった、のだ、、、、、、、私は、５７歳だ、、、、私が、８１０歳の頃の記憶では、正月に、初参りに行った時、、、、、、、真実は歴史の闇に消えて行く、のだ、のだ、のだ、のだ、、、、、、、、、これで良い、、、、、、エドコンでの、撮影が、、多い、、やに、、感じる、、、、、下の左は、のアマゾニア、以前のファイヤハウスの所か？ |
| **Text b)** |
| たまにはにぎやかな場所にも行きたい。でも、私は長い時間スーパーなどに居ることが出来ない。カートを持ってぶつかりながら狭いところを通るたびに「すみません通してください」と何度その言葉を言わなければならないか。言葉を発することは私にとって辛い。身内ともめったに話しなどしない。しても前は電話を切られた。だから兄のところに相談に行った。私には人と話すのは限度がある。確かにもっと悪い人もいるだろう。でも、それを例に挙げて私に行っても私は自分しか見えていない。話しを聞いても分からない。最近病院へ行くのも嫌になっている。私も記憶障害があるのでその場で受け答えが出来ないから余計に医らつくのだと思う。いいたいことの半分も話が出来ない、忘れてしまっているから |

Highlights shown in red denote character or character-sequence usages unique to the text's author.

Text a) is one of the two texts in the SA comparison that yielded the highest SA LR ($\log_{10}$ LR = 28.03885). Its counterpart cannot be shown in Table 17 because the BCCWJ database does not publicly link authorship across separate texts. The author's most distinctive feature is the frequent, sequential use of the Japanese comma '、', which recurs throughout the passage. Additionally, individual characters or character sequences are repeatedly employed (highlighted in red in Table 17). These non-standard orthographic choices, together with characteristic vocabulary and grammatical constructions, create a distinctly colloquial register. The same features are evident in the paired text, and these consistent idiosyncrasies across both passages naturally produces a high LR.

Texts a) and b) in Table 17—authored by different individuals—yielded the greatest negative DA LR ($\log_{10}$ LR = –9.90240). In contrast to Text a)'s distinctive style, Text b) adheres to standard Japanese orthography and employs a neutral, diary-like register. A straightforward visual comparison between the two excerpts reveals marked stylistic differences, which accounts for the pronounced negative $\log_{10}$ LR.

## 8 SUMMARY AND DISCUSSIONS

The following is a summary of the present study regarding fusion.

Among the stylometric-feature systems:

- The optimal fusion combines Character-bigrams (System 1), Comma-bigrams (System 3) and POS-bigrams (System 4), achieving a $C_{llr}$ = 0.43964.

Individually, the Character-bigram system (System 1) performs best, followed by the POS-bigram system (System 4), with the Comma-bigram system (System 3) ranking fourth. This demonstrates that simply increasing the number of features does not guarantee improved performance (Ishihara 2014, 2017). Effective fusion depends on the complementarity of the information each feature encodes: the most successful fused systems combine features that capture different dimensions of authorship, not necessarily the top single-feature performers.

For example, the most effective two-feature fusion ($C_{llr}$ = 0.44280) combines Character-bigrams with Comma-bigrams. Although both features draw on character bigrams, they overlap in only nineteen instances and otherwise consist of largely distinct sets—thereby supplying complementary stylistic cues. By contrast, POS-bigrams encode syntactic patterns rather than raw character sequences. This difference in linguistic domain underpins the strong complementarity of POS-bigrams to both Character- and Comma-bigrams, explaining their joint effectiveness in fusion.

Among the five stylometric features, Function-word unigrams are the only word-level N-grams and were therefore expected to complement the other features. In practice, however, they showed limited complementarity. For example, the best two-feature fusion involving Function-word unigrams—paired with Character-bigrams—yielded a relatively modest $C_{llr}$ of 0.47106.

For fusion, the feature-vector dimensions were fixed at each system's optimal setting: 500 for Character-bigrams (System 1), 125 for Function-word unigrams (System 2), 40 for Comma-bigrams (System 3) and 500 for POS-bigrams (System 4). Feature elements—such as adjacent bigrams—are not statistically independent (for example, the bigram ('し', 'た') often co-occurs with the bigram ('た', '。'), introducing redundancy into the vectors. The robust system should maintain or enhance its accuracy even if redundant features are removed. Techniques for identifying and discarding such redundancies—commonly referred to as feature selection or feature reduction—are widespread in NLP (Dey Sarkar et al. 2014; Houvardas and Stamatatos 2006; Villar-Rodriguez et al. 2016). Assessing their impact on system performance warrants future investigation. Reducing vector dimensionality also enhances system stability, since higher dimensions demand exponentially more training data to achieve reliable estimates (Silverman 1986).

Amongst the embedding-based systems:

- The fusion of the two embedding-based systems: one is based on word-embedding and the other is on character-embedding, resulted in a better performance ($C_{llr}$ = 0.36186) than each embedding-based system alone and the optimal fused stylometric-feature system ($C_{llr}$ = 0.43964).
- The magnitude of the consistent-with-fact LRs is conservative; See Figure 6b).

The strong performance of fusing the two embedding-based systems is unsurprising, since one uses word embeddings and the other character embeddings—each capturing different aspects of the text. However, the character-based RoBERTa was constrained by a max_seq_length of 512 tokens, preventing it from processing entire documents and likely limiting its effectiveness. Further research should therefore investigate optimized character-embedding methods that accommodate longer sequences.

The embedding-based systems produce cosine similarity scores, which are inherently bounded between –1 and 1. This limited score range can constrain the magnitude of LRs derived from fusing embedding-based outputs. Cosine distance was employed in this study as the scoring function for the practical reasons outlined in Section 5.1.2. From a theoretical standpoint, however, a similarity-and-typicality approach—such as PLDA (Prince and Elder 2007)—is more appropriate for LR-based systems (cf. Ishihara et al. 2022). Incorporating PLDA should therefore be considered a valuable extension of this work.

Between the stylometric-feature and embedding-based systems:

- The fusion of Systems 1 (Character-bigrams), 4 (Comma-bigrams), 5 (Character types), 6 (Word embeddings) and 7 (Character embeddings) yields the best overall performance ($C_{llr}$ = 0.32484).
- This fused system retains excellent calibration while simultaneously
    - Enhancing the magnitude of consistent-with-fact LRs;
    - Decreasing the magnitude of contrary-to-fact LRs and
    - Improving overall discriminability.

As shown in Table 16, fusing Systems 1, 4, 5, 6 and 7 yields the lowest $C_{llr}$ (0.32484), only marginally better than fusing all seven systems ($C_{llr}$ = 0.32855). In fact, the top-performing fusions—from three to seven systems—all fall within a narrow band around 0.32, indicating that even the three-system fusion (Systems 1, 6 and 7) approaches optimal performance. This pattern suggests that the remaining systems, while not contradictory, do not supply substantially new authorship information.

As noted above, fusion yielded higher consistent-with-fact LRs and lower contrary-to-fact LRs. However, the maximum SA $\log_{10}$ LR of 28.03885 produced by the best-performing fusion is implausibly large—even exceeding typical values in DNA typing (Willis et al. 2015)—indicating instability in the system. This extreme value appears to originate from the fused stylometric-feature system (see Figure 6a). Further investigation is needed to identify the cause of these outliers and to explore the use of shrinkage techniques to temper excessively large LRs (Morrison and Poh 2018; Vergeer et al. 2016).

## 9 FUTURE WORK

This study used approximately 1,000-character excerpts from Japanese blogs to demonstrate the effectiveness of logistic regression fusion between stylometric-feature and embedding-based systems in LR-based FTC framework. However, blogging is just one genre, and writing style varies markedly across platforms and contexts (e.g. email, social media, websites), influenced by factors such as topic, emotion, interlocutor and formality. To fully assess fusion's utility, future research must include diverse text types that span these conditions. In real casework, mismatches—such as comparing an email sample to social-media text—are inevitable and degrade performance. Empirical validation under such realistic mismatched conditions is therefore essential to build robust, reliable FTC systems (Morrison et al. 2025; Morrison et al. 2021a).

The LR framework has started being recognized as a logically and legally appropriate basis for FTC (Grant 2022), yet its application has been largely confined to English (Ishihara 2017, 2021, 2023; Ishihara and Carne 2022; Ishihara et al. 2024; Nini 2023; Nini et al. 2026). To ensure its broad applicability, LR-based FTC systems must undergo rigorous empirical validation in multiple languages. Writing systems differ substantially, the most discriminative features are often language-specific, and the availability and performance of LLMs can vary greatly between languages. Addressing these factors is essential for developing empirically validated and reliable FTC methods that are applicable across a wide variety of languages.

## APPENDICES

### A.1 Example Yahoo! Blog Excerpt

An excerpt from the text of sample ID OY03_01841.

さて、ＧＷの二日目をちゃちゃっとＵＰして、明日からに備えようｗ。なんか一日をＵＰするのに一週間かけてすみませんね（＾▽＾）／。お風呂にはいった後は飯を食べようと、滑川に行きましたが、これまた、良い食堂がありませんでしたので、ちょっと魚津まで足をのばしてみました。魚津は漁港すからね、寿司をたべることに１００円ですがｗ１５０円さらもありましたけどねｖ（＊'－＾＊）ｂぶいまずはほたるイカでしょ、初めて回転寿司でたべましたヾ（@⌒¬⌒@）ノ、１００円とは思えないウマさでしたね、さすが、富山です刺身もふつうにうまかったです金色の皿は１５０円よ（＾▽＾）／。そうそう、ここ設備はハイテク注文はすべてタッチパネルでＯＫ追加注文は上の高速新幹線ではこんでくれますｗ。レールがなかったので、リニアかもしれまねんね！！最後にくったのがなんだっけ、これうまかったっす嫁さんはしめにうどんをいやはや、魚津の回転寿司は最高でした、またいきたいね。回転寿司かいおう魚津店［すし］Ｙａｈｏｏ！グルメ地図を拡大して表示住所富山県魚津市上村木４１３最寄り駅魚津Ｙａｈｏｏ！グルメでこの店舗のクチコミを見る２００８年６月２９日時点の情報です。ＩＤ０００７７７８５８８お店情報を自分のブログにもはりつけようさて、おなかいっぱい食った後は就寝場所へｗ。滑川の道の駅のにてウェーブパーク滑川です！！ほたるイカのライトが明日への希望をもりあげてくれました、おやすみなさい。本日の工程写真撮影５２６枚移動距離１００Ｋｍ？立山駐車場美女平室堂雪の大谷美女平立山カルデラ博物館称名滝温泉魚津回転寿司滑川道の駅さすがに車泊につかれたと思い出す３日目の夜でした（＾＾；；さて、新婚生活でたのしみな食卓公開シリーズです、まぁ新婚のうち、又、めずらしいものが出てきたときには写真にでもとって、おしげもなく公開しますよ、もっとほうがいいんじゃない、みたいな先輩夫婦の助言たのんます。上の２行がコピーなのは目をつぶってください、気づいてるでしょうが。写真は２００８年のよｗ愛の料理開幕です？ついに弁当キタ（ﾟ∀ﾟ）！！冷凍か否かは（￣¬￣）クチチャックです。凍った飯をつかったら絶対焼き飯なんですよね。おにぎり作ってほしくてダイソーで型を自ら買ってきました！！餃子チャーハンですなんか太りそうな弁当ですね。嫁のカイワレ巻きハムは絶品す。この辺で、朝いそがしいときなので、めったに撮影してませんぜ、遅刻するぜ！！

### A.2 Supplementary Data Files

- List of the 500 most frequently used character bigrams
- List of the 500 most frequently used function words
- List of the 500 most frequently used POS bigrams
- List of the 200 most frequently used bigrams of the comma and the preceding character